\documentclass[conference]{IEEEtran}
\newif\ifblind
\blindfalse
\usepackage{cite}
\usepackage{amsmath,amssymb,amsfonts}
\usepackage{graphicx}
\usepackage{capt-of} % Mixed table/figure float; retain IEEEtran caption styling.
\usepackage{booktabs}
\usepackage{multirow}
\newcommand{\cmark}{\ding{51}}
\newcommand{\xmark}{\ding{55}}
\usepackage[table]{xcolor}      % [table] -> colortbl, enables \rowcolor (shaded "ours" rows in tab:main)
\definecolor{ourrow}{gray}{0.92}  % light grey for our-method rows
\usepackage{url}
\usepackage{tabularx}
\usepackage{makecell}
\usepackage{pifont}
\usepackage{tikz}
\usetikzlibrary{positioning, arrows.meta, fit, calc}
\usepackage[hidelinks]{hyperref}
\usepackage[capitalize,noabbrev]{cleveref}

\begin{document}

% PLACEHOLDER TITLE -- drops "Benchmark" and the method claim; sharpen once the method lands.
\title{CytoSPM: Open-Vocabulary Cytopathology Detection with Structured Prompt Bank}

\ifblind
  \author{\IEEEauthorblockN{Anonymous Author(s)}
  \IEEEauthorblockA{Affiliation withheld for double-blind review}}
\else
  \author{
    \IEEEauthorblockN{
      Wenjie Li\textsuperscript{1,*},
      Zishan Xu\textsuperscript{2,*},
      Jinyang Huang\textsuperscript{1},\\
      Zhengxin Nie\textsuperscript{1},
      Shichao Kan\textsuperscript{1}, and
      Yixiong Liang\textsuperscript{1,\ensuremath{\dagger}}}
    \IEEEauthorblockA{
      \textsuperscript{1}Central South University 
      \textsuperscript{2}Shanghai Jiao Tong University\\
          \textsuperscript{*}Equal contribution.
      \textsuperscript{\ensuremath{\dagger}}Corresponding author.}
      {\small\texttt{254711081@csu.edu.cn, zishanxu@sjtu.edu.cn, yxliang@csu.edu.cn}}\\
  
  }
\fi

\maketitle

\begin{abstract}
Cytopathology detection requires open-vocabulary recognition because cellular categories are fine-grained, long-tailed, and continuously evolving across different organ systems. However, existing cytology detectors are mostly single-domain and closed-set, and there is still no unified benchmark for evaluating open-vocabulary cytopathology detection. We present \textbf{PentaCyto}, a multi-domain benchmark covering cervical, urinary, respiratory, serous fluid, and thyroid cytology, with 24 base categories and 9 held-out novel categories. Each category is associated with structured cytomorphology prompts that describe diagnostic morphological attributes and provide clinically grounded textual knowledge. We further propose \textbf{CytoSPM}, an efficient detector based on a decoupled two-stage design. It first extracts reusable class-agnostic visual representations, and then performs class-aware structural prompt matching with class names and cytomorphology prompts. On PentaCyto, CytoSPM outperforms existing methods in novel-category detection and open-vocabulary detection while maintaining efficient inference.
\end{abstract}


\section{Introduction}
\label{sec:intro}

Cytopathology is an important tool for clinical diagnosis and cancer
screening, where automatically localising and classifying abnormal cells
is a core component of computer-aided cytological screening. Although
deep learning has advanced object detection~\cite{ren2015faster,he2017mask,redmon2016you,zhao2024detrs},
segmentation~\cite{guo2026decoupling,lu2026zeroforgetting,xu2026videosegr1},
and multimodal understanding~\cite{du2026ledgermind,guo2025octopusagenticmultimodalreasoning},
most existing cytopathology detectors remain limited to the closed-set
paradigm and can only recognise categories predefined during training.
This assumption does not match the fine-grained, long-tailed and
continuously evolving label space of cytopathology: different organ
systems follow distinct diagnostic criteria, many categories are
separated only by subtle morphological differences, and rare subtypes or
newly described entities are continually incorporated into diagnostic
practice. Therefore, cytopathology detection requires open-vocabulary
capability to recognise unseen and long-tail cellular categories.

Open-vocabulary object detection addresses the category limitation of
closed-set detectors. Through vision--language alignment or region-level
vision--language pre-training~\cite{radford2021learning,jia2021scaling}, open-vocabulary detectors can recognise
and localise categories specified by textual prompts, including those
unseen during detector training. Recent methods such as
ViLD~\cite{gu2021open}, RegionCLIP~\cite{zhong2022regionclip},
OWL-ViT~\cite{minderer2022owlvit}, GLIP~\cite{li2022grounded},
GroundingDINO~\cite{liu2024grounding}, YOLO-World~\cite{cheng2024yolo}
and WeDetect~\cite{fu2026wedetect} have achieved strong progress on
natural-image open-vocabulary detection benchmarks. This paradigm offers
a promising direction for cytopathology, since unseen cellular categories
can in principle be queried through textual prompts. However, despite
its success in natural images, open-vocabulary detection remains largely
under-explored in cytopathology.
\begin{figure}[t]
  \centering
  \includegraphics[width=\columnwidth]{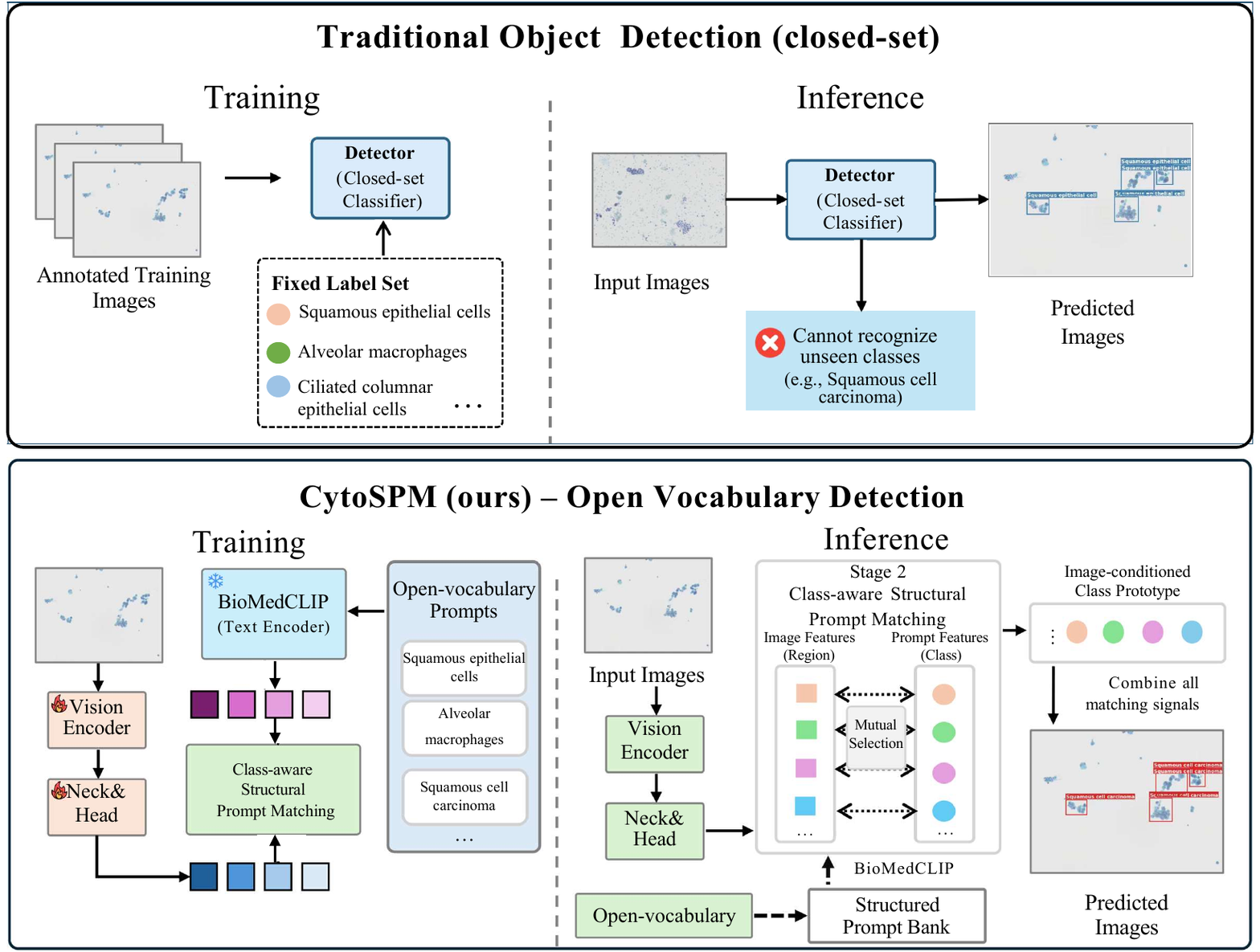}
  \caption{
  \textbf{Comparison of traditional closed-set detection and CytoSPM.}
  Traditional object detection methods rely on a fixed label set and can only
  recognize cell categories predefined during training, making it difficult
  to generalize to unseen, rare, or newly defined cytopathology classes.
  In contrast, CytoSPM addresses open-vocabulary cytopathology detection with
  a decoupled two-stage design. It first extracts reusable class-agnostic
  visual representations from cytology images, and then performs class-aware
  structural prompt matching with cytomorphology prompts. By combining
  class-name, structural prototype, and fine-grained attribute matching,
  CytoSPM enables text-prompt-driven detection of both seen and unseen
  cytology cell categories.
  }
  \label{fig:intro}
\end{figure}
Specifically, applying open-vocabulary detection to cytopathology faces
two key challenges. First, there is no public benchmark for
open-vocabulary cytopathology detection. Existing cytology datasets are
mostly single-domain and closed-set, making it difficult to evaluate
cross-domain recognition of novel cellular categories. Meanwhile,
cytopathology lacks large-scale, diverse and well-annotated image--text
detection data, which limits the ability of models to learn reliable
medical vision--language associations. As a result, open-vocabulary cell
detection still lacks a unified and reproducible evaluation foundation.

Second, efficient open-vocabulary detection methods for high-throughput
cytopathology screening are still insufficient. Existing high-performing
open-vocabulary detectors often rely on deep cross-modal fusion to
strengthen vision--language alignment. Although effective, such fusion
modules introduce substantial computational overhead and make visual
features dependent on specific text queries, preventing feature sharing
across different categories or prompts and reducing inference efficiency
in large-vocabulary settings. In contrast, efficient detectors such as
YOLO-World\cite{cheng2024yolo} and WeDetect\cite{fu2026wedetect} adopt lighter vision--language matching
paradigms: YOLO-World\cite{cheng2024yolo} combines a lightweight YOLO framework with a CLIP\cite{radford2021learning}
text encoder and uses a re-parameterisable vision--language path
aggregation module for real-time detection, while WeDetect\cite{fu2026wedetect} adopts a
simpler dual-tower architecture without cross-modal fusion layers in the
detection neck, performing classification by dot products between image
grid features and class text embeddings. Nevertheless, these methods are
mainly designed for natural images and do not directly address the
long-tail distribution, strong domain shift, small object scale,
morphological similarity and fine-grained category distinctions in
cytopathology. Therefore, directly transferring general open-vocabulary
detectors remains insufficient for jointly achieving open-category
recognition, long-tail generalisation and high-throughput screening
efficiency.

To address these challenges, we first construct \textbf{PentaCyto}, a
new multi-domain benchmark for open-vocabulary cytopathology detection.
PentaCyto covers five clinically distinct cytopathology domains,
including cervical, urinary, respiratory, serous fluid and thyroid
cytology. It contains 24 base categories drawn from corresponding
national reporting standards, together with a category-disjoint pool of
clinically significant novel malignancies for open-vocabulary evaluation.
More importantly, each category is associated with a structured
cytomorphology prompt that summarises the diagnostic attributes used by
pathologists to recognise the corresponding cells, providing
clinically-grounded textual knowledge for open-vocabulary detectors.

As shown in Figure~\ref{fig:intro}, we further propose \textbf{CytoSPM}, an efficient open-vocabulary
cytopathology detection framework based on a decoupled two-stage design.
The first stage is a \textbf{class-agnostic visual feature extraction stage}: the
model generates general region-level visual representations from the
input image using only the visual branch, without introducing class
names, textual prompts or cross-modal fusion. These visual features can
therefore be reused across different categories and prompts while
preserving real-time inference efficiency. The second stage is a
\textbf{class-aware object detection stage}: the model uses class names and a
structured morphology-attribute prompt bank to perform class-aware
matching over the general visual regions produced by the first stage.
Specifically, for each category, we construct a prompt bank of
cytomorphological attributes and use a mutual-selection mechanism to
dynamically select category-relevant visual evidence and structural
prompts. The selected prompts are then aggregated into an
image-conditioned class prototype. Finally, classification is performed
by combining three types of matching signals: class-name, structural
prototype and fine-grained attribute matching. In this way, CytoSPM
decouples class-agnostic visual representation extraction from
class-aware structural detection, avoiding heavy cross-modal fusion while
maintaining efficient inference and improving fine-grained and long-tail
cell recognition.

We evaluate the proposed framework on PentaCyto across all five
cytopathology domains. The results show that our method preserves strong
base-category detection performance, improves open-vocabulary transfer
to held-out novel cellular categories, and maintains efficient inference
suitable for high-throughput cytopathology screening. Our contributions are summarized below.

\begin{itemize}
\item We introduce \textbf{PentaCyto}, the first multi-domain benchmark for open-vocabulary cytopathology detection. It covers five cytology domains, including cervical, urinary, respiratory, serous fluid, and thyroid cytology, with 24 base categories and a category-disjoint novel pool, providing a unified platform for cross-domain generalization and novel-category detection.

\item We propose \textbf{CytoSPM}, an efficient open-vocabulary cytopathology detection framework with a decoupled two-stage design. It first extracts class-agnostic region-level visual features, then selects category-relevant visual evidence and morphological attributes through structural prompt matching, and finally combines class-name, structural prototype, and fine-grained attribute matching for classification.

\item We conduct systematic experiments on PentaCyto. CytoSPM achieves \textbf{SOTA} performance in both overall detection and novel-category detection, while maintaining strong base-category performance and efficient inference.
\end{itemize}

\begin{table*}[t]
  \centering
\caption{
\textbf{Statistics of the PentaCyto dataset.}
PentaCyto covers five cytopathology domains with domain-specific reporting standards.
For each domain, we report the number of categories, whole-slide images (WSIs), images,
and bounding-box annotations under the base and novel splits. Dashes indicate that no
novel categories are defined for the corresponding domain.
}
  \label{tab:dataset_statistics}
  \setlength{\tabcolsep}{5pt}
  \newcommand{\NA}{\multicolumn{1}{c}{--}}
  \begin{tabular}{l l c r r r c r r r}
    \toprule
    & & \multicolumn{4}{c}{\textbf{Base}} & \multicolumn{4}{c}{\textbf{Novel}} \\
    \cmidrule(lr){3-6} \cmidrule(lr){7-10}
    Domain & Standard & \#\,Cat. & \#\,WSI & \#\,Images & \#\,Boxes & \#\,Cat. & \#\,WSI & \#\,Images & \#\,Boxes \\
    \midrule
    Cervical & TBS      & 9 & 1{,}588& 34{,}305 & 123{,}474 & \NA & \NA & \NA & \NA \\
    Urinary                                  & Paris    & 3 & 220                  & 25{,}720 &   8{,}410 & \NA & \NA & \NA & \NA \\
    Respiratory                              & PSC      & 6 & 162                  & 50{,}102 & 611{,}320 & 3   & 20  &  2{,}593 &  3{,}517 \\
    Serous fluid                             & TIS      & 1 & 10                   &  4{,}187 &  19{,}705 & 3   & 16  &  6{,}213 & 17{,}445 \\
    Thyroid                                  & Bethesda & 5 & 261                  & 40{,}719 & 149{,}172 & 3   & 15  &  2{,}656 &  8{,}250 \\
    \midrule
    \textbf{Total}                           &          & \textbf{24} & \textbf{2{,}241} & \textbf{155{,}033} & \textbf{912{,}081} & \textbf{9} & \textbf{51} & \textbf{11{,}462} & \textbf{29{,}212} \\
    \bottomrule
  \end{tabular}
\end{table*}

\begin{figure*}[t]
  \centering
  \includegraphics[width=0.9\textwidth]{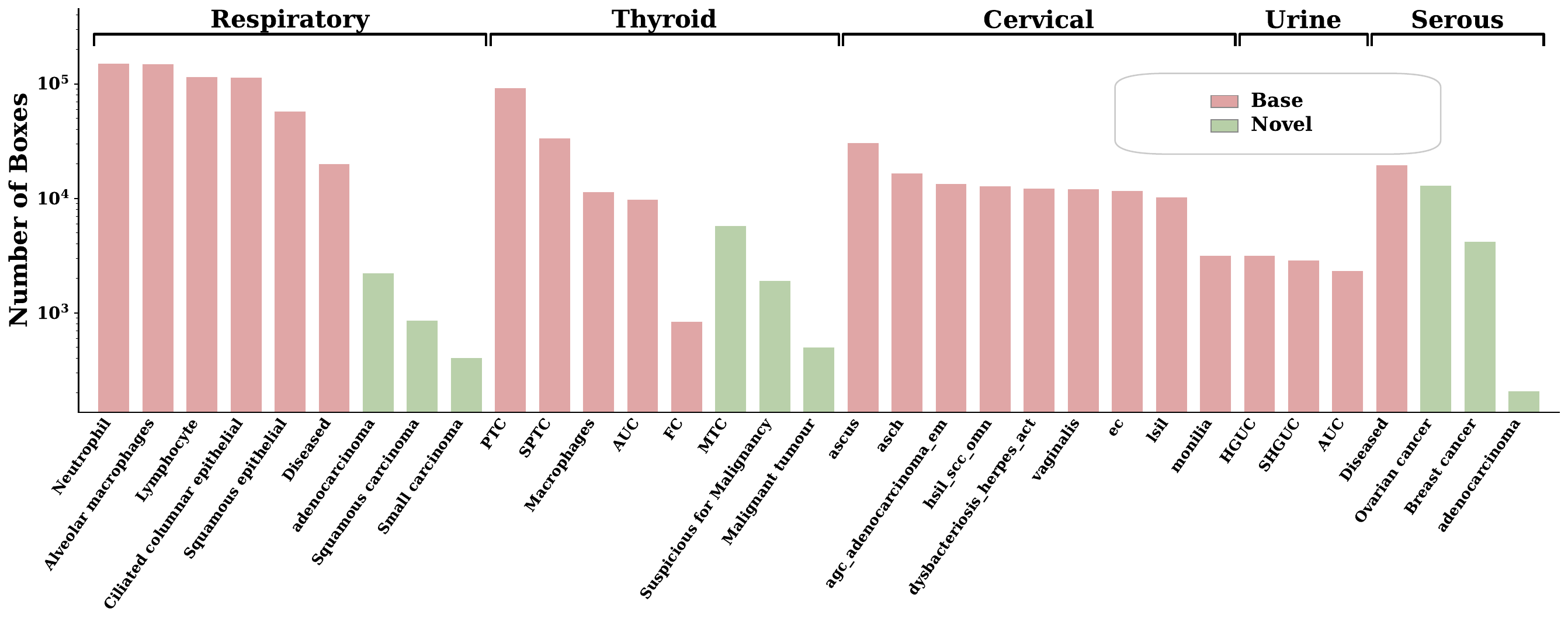}
  \vspace{-0.9em}
  \caption{The bar chart presents the bounding-box counts of each category across five cytology domains.}
  \label{fig:classdist}
\end{figure*}
\section{Related Work}
\label{sec:related}
\subsection{Object Detection in Medical Images}

Object detection plays a central role in medical image analysis, enabling the localization and recognition of tumors, lesions, cells, and other diagnostically relevant abnormalities. Recent deep learning-based detectors, originally developed for natural images, have been widely adapted to medical imaging scenarios. For example, BGF-YOLO\cite{kang2024bgf} and SOCR-YOLO\cite{liu2024socr} introduce YOLO-based architectural improvements for brain tumor detection and lesion detection, respectively. In cell-related tasks, HERO \cite{jiang2024holistic} further focuses on fine-grained instance discrimination in cervical cell detection, promoting the development of cell-level object detection. However, these methods are typically built upon a closed-set detection paradigm, where models can only recognize cell categories predefined during training. Meanwhile, existing cell detection studies are mostly limited to a single organ or a single data domain, such as cervical cytology images, and there remains a lack of a unified multi-domain benchmark for systematically evaluating open-vocabulary cell detection. As a result, existing methods are insufficient for assessing model generalization across anatomical sites, diagnostic systems, and unseen categories, which further limits the development of open-vocabulary cytopathology detection.

\subsection{Open-Vocabulary Object Detection}

Open-vocabulary object detection aims to detect both known and novel categories by aligning visual regions with textual concepts. In recent years, substantial progress has been made in natural image domains, where representative methods usually formulate object detection as region-text matching or phrase grounding. GLIP~\cite{li2022grounded} unifies object detection and phrase grounding through region-word contrastive pre-training, while Grounding DINO~\cite{liu2024grounding} further incorporates grounded pre-training into detection Transformers and enhances vision-language alignment through cross-modal fusion. Meanwhile, efficient detectors such as YOLO-World~\cite{cheng2024yolo} combine YOLO-style architectures with vision-language representations, achieving open-vocabulary detection while maintaining real-time inference efficiency and practical downstream deployment. Recently, MedROV~\cite{sheikh2026medrov} further introduces open-vocabulary detection into medical imaging by constructing a large-scale multi-modal medical detection dataset and incorporating medical vision-language foundation models to improve the detection of both known and unseen medical structures. However, existing medical open-vocabulary detection studies still pay limited attention to fine-grained, long-tailed, and cross-domain cytopathology scenarios. To fill this gap, we introduce an efficient open-vocabulary cytopathology detection framework that breaks the limitation of predefined closed-set categories and generalizes to unseen categories in real-world cytology screening while maintaining strong detection performance and inference efficiency.

\begin{figure*}[t]
  \centering
  \includegraphics[width=1.5\columnwidth]{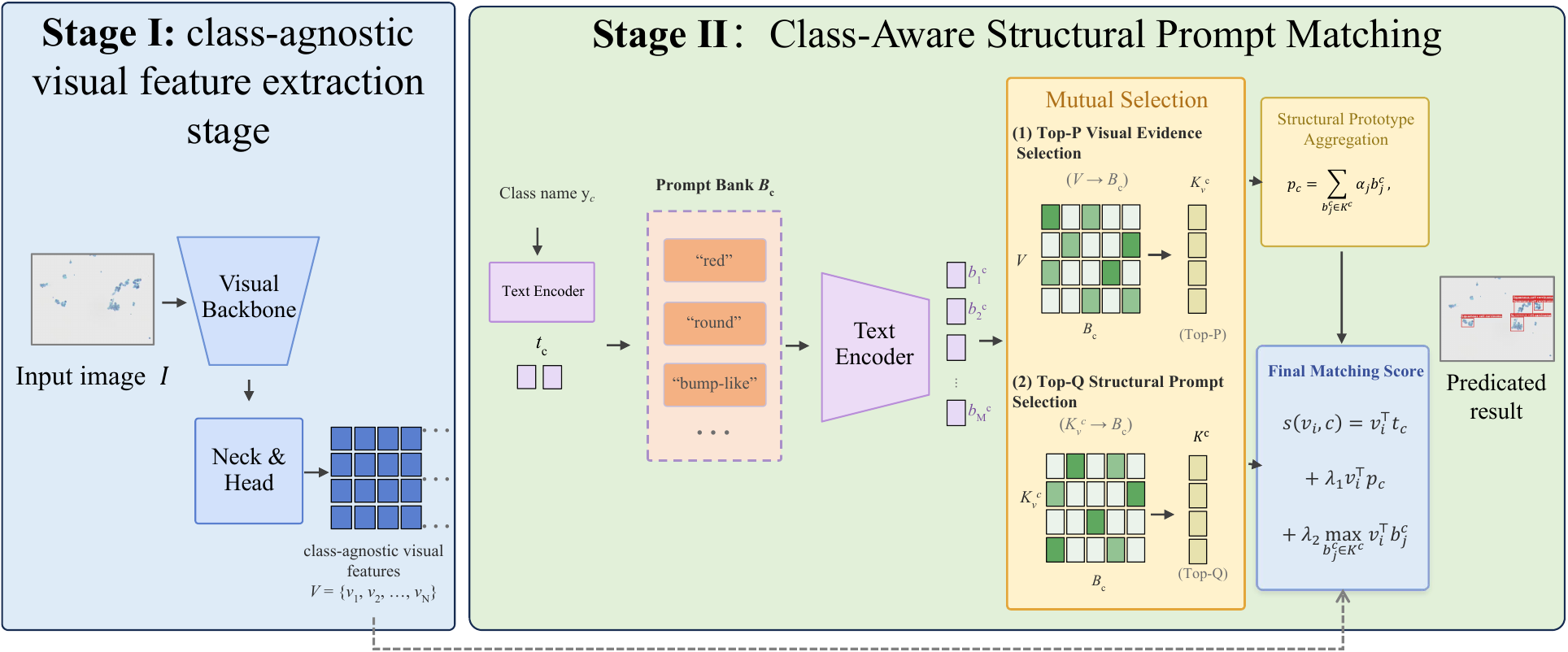}
 \caption{
\textbf{Overview of CytoSPM.}
CytoSPM is an efficient open-vocabulary cytopathology detection framework based on a decoupled two-stage design.
In Stage I, the visual branch extracts reusable class-agnostic region-level representations from the input cytology image without introducing class names, textual prompts, or cross-modal fusion.
In Stage II, class names and class-specific cytomorphology prompt banks are encoded by a frozen text encoder, and class-aware structural prompt matching is performed over the visual regions.
Through mutual selection, CytoSPM dynamically selects category-relevant visual evidence and visually supported morphology prompts, which are aggregated into an image-conditioned structural prototype.
The final classification score combines class-name matching, structural prototype matching, and fine-grained attribute matching, enabling efficient recognition of both seen and unseen cytopathology categories.
}
  \label{fig:method}
\end{figure*}
\section{PentaCyto Dataset}

Cytopathology data naturally exhibit substantial domain gaps. 
Different cytology domains not only follow distinct clinical reporting systems, 
but also present domain-specific cellular morphologies, background patterns, 
and category distributions. Existing cytopathology detection datasets are 
mostly limited to a single domain or a fixed category set, making it difficult 
to evaluate model generalization across domains, fine-grained categories, and 
open-vocabulary settings. As summarized in Table~\ref{tab:related_datasets_methods},
existing related works cover medical open-vocabulary detection, cytology
foundation models, and single-domain cell detection, but none of them provides
a unified benchmark for multi-domain cell-level open-vocabulary cytopathology
detection. To this end, we construct PentaCyto, a 
multi-domain cytology cell detection dataset with an open-vocabulary evaluation 
protocol. PentaCyto covers five cytology domains, including respiratory, 
cervical, thyroid, urinary, and serous fluid cytology, and defines its 
categories according to the corresponding clinical reporting systems, namely 
PSC, TBS, Bethesda, Paris, and TIS. To ensure annotation quality, we invited 
eight experts with medical backgrounds to participate in bounding-box annotation 
and category verification. After the initial annotation stage, cross-review was 
conducted among experts to reduce missed annotations, incorrect labels, and 
potential confusion among fine-grained categories.

As shown in Table~\ref{tab:dataset_statistics}, PentaCyto contains 
33 diagnostic categories, including 24 base diagnostic categories for training, 
validation, and base testing, and 9 novel diagnostic categories that are 
completely unseen during training and are used only for open-vocabulary evaluation. For the base split, PentaCyto consists of 2,241 WSI images, 155,033 cytology images, and 912,081 
diagnostic cell bounding-box annotations. To evaluate generalization to unseen 
categories, we further construct an independent novel pool, which contains 
9 novel diagnostic categories, 51 WSIs, 11,462 images, and 29,212 cell 
bounding-box annotations. All novel categories and their corresponding images 
are excluded from training. This design makes PentaCyto a challenging benchmark 
for long-tailed category distributions, fine-grained morphological recognition, 
and cross-domain generalization in cytopathology detection. Figure~\ref{fig:classdist} further shows the bounding-box distribution of each category across the five cytology domains.

\begin{table}[t]
\centering
\caption{
Comparison with related cytology and medical open-vocabulary detection works. Multi indicates whether multiple medical or cytology domains are covered; Cls. denotes classification, and Det. denotes detection, which requires both cell localization and category recognition.
}
\label{tab:related_datasets_methods}
\scriptsize
\setlength{\tabcolsep}{2.5pt}
\renewcommand{\arraystretch}{1.12}
\begin{tabular}{p{1.35cm} c p{4.25cm} p{1.25cm}}
\toprule
\textbf{Name} & \textbf{Multi} & \textbf{Domain} & \textbf{Task} \\
\midrule
MedROV~\cite{sheikh2026medrov} & \cmark &  CT, MRI, ultrasound, X-ray,  histopathology, colonoscopy, microscopy, dermoscopy, and fundoscopy  & Det. \\
CytoFM~\cite{ivezic2025cytofm} & \cmark &  Breast, cervical, and thyroid & Cls. \\
CytoCrowd~\cite{si2026cytocrowd} & \cmark & Endocervical, metaplastic cells, etc. & Det. / Cls. \\
CytoSAE~\cite{dasdelen2025cytosae} & \xmark & Hematology cytology & Det. \\
HERO~\cite{jiang2024holistic} & \xmark & Cervical cytology & Det. \\
\bottomrule
\end{tabular}
\end{table}

% \begin{figure}[t]
%   \centering
%   \includegraphics[width=\columnwidth]{figs/fig_boxsize.pdf}
%   \caption{Bounding-box size distribution per organ (violin of
%   $\sqrt{\mathrm{area}}$ over all base boxes; white dot = median; dashed line
%   = the COCO small-object threshold $\sqrt{\mathrm{area}}<32$\,px). The
%   fraction of boxes below 32\,px is printed above each organ. Cytology
%   targets are predominantly small (respiratory $85\%$, urine $84\%$), so
%   localisation at high IoU is intrinsically hard; we therefore report
%   AP$_{50}$ alongside the stricter AP$_{.5:.95}$.}
%   \label{fig:boxsize}
% \end{figure}

\begin{table*}[t]
  \centering
  \small
  \caption{\textbf{Comparison on the PentaCyto. } YOLOE (text), (visual), and (mixed) use textual prompts, visual prompts, and both prompts, respectively. The best and second-best results are shown in \textbf{bold} and \underline{underlined}.}
  \label{tab:main}
  \setlength{\tabcolsep}{4.8pt}
  \begin{tabular}{l l r r rrr rrr r}
    \toprule
    & & & & \multicolumn{3}{c}{Base} & \multicolumn{3}{c}{Novel} & \\
    \cmidrule(lr){5-7}\cmidrule(lr){8-10}
    Method & Backbone & \#Params & FPS
      & AP & AP$_{50}$ & AP$_{75}$
      & AP & AP$_{50}$ & AP$_{75}$ & AP$_{\mathrm{AVG}}$ \\
    \midrule
    GLIP~\cite{li2022grounded}
      & Swin-T & 231.8M & \phantom{0}7.1
      & 28.90 & 43.50 & 32.70
      & 12.10 & 16.90 & 14.10 & 20.50 \\

    Grounding DINO~\cite{liu2024grounding}
      & Swin-T & 173.5M & \phantom{0}6.2
      & 30.40 & 47.50 & 33.80
      & \phantom{0}2.80 & \phantom{0}4.30 & \phantom{0}3.30 & 16.60 \\

    OV-DEIM~\cite{wang2026ov}
      & DINOv3-ViT-T & 11.4M & 23.2
      & 31.60 & 45.50 & 37.20
      & 12.70 & 17.00 & 14.80 & 22.15 \\

    YOLOE (text)~\cite{wang2025yoloe}
      & YOLOv8-L & 51.2M & \textbf{75.3}
      & \underline{33.60} & 48.30 & \textbf{39.10}
      & \phantom{0}6.40 & \phantom{0}8.80 & \phantom{0}7.40 & 20.00 \\

    YOLOE (visual)~\cite{wang2025yoloe}
      & YOLOv8-L & 51.2M & \textbf{75.3}
      & 29.70 & 43.10 & 34.30
      & 12.50 & 16.90 & 14.90 & 21.10 \\

    YOLOE (mixed)~\cite{wang2025yoloe}
      & YOLOv8-L & 51.2M & \textbf{75.3}
      & 32.80 & 47.40 & 38.00
      & 11.50 & 15.60 & 13.60 & 22.15 \\

    YOLO-World-L~\cite{cheng2024yolo}
      & YOLOv8-L & 110.5M & 25.7
      & 32.40 & 48.20 & 37.50
      & \underline{13.70} & 18.30 & \underline{15.80} & 23.05 \\

    LLMDet~\cite{fu2025llmdet}
      & Swin-T & 173.0M & \phantom{0}6.6
      & \textbf{33.80} & 50.90 & \underline{38.40}
      & \phantom{0}7.60 & 10.90 & \phantom{0}8.90 & 20.70 \\
      
    MedROV~\cite{sheikh2026medrov}
      & YOLOv8-L & 51.6M & \underline{52.2}
      & 27.96 & 39.25 & 32.66
      & 3.62 & 4.76 & 4.03 & 15.79 \\
      
    WeDetect~\cite{fu2026wedetect}
      & ConvNeXt-T & 37.9M & 34.3
      & 33.37 & \underline{51.43} & 38.15
      & 13.30 & \underline{18.98} & 15.34 & \underline{23.34} \\

    \midrule
    \textbf{CytoSPM (ours)}
      & ConvNeXt-T & 37.9M & 25.8
      & 33.55 & \textbf{52.20} & 38.17
      & \textbf{15.00} & \textbf{21.00} & \textbf{17.20} & \textbf{24.27} \\
    \bottomrule
  \end{tabular}
\end{table*}

\begin{figure*}[t]
  \centering
  \includegraphics[width=0.71\textwidth]{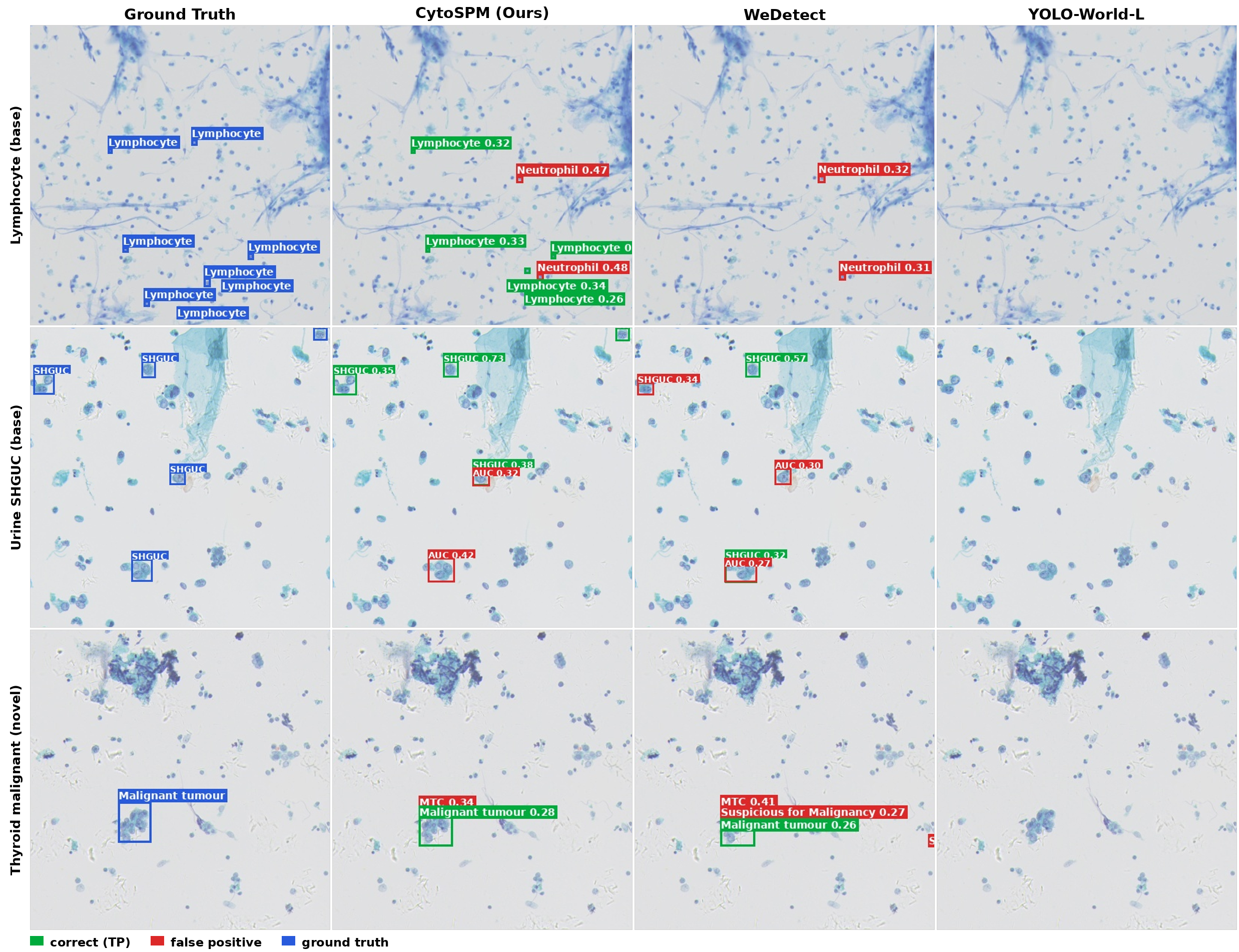}
  \caption{
\textbf{Qualitative comparison on PentaCyto.} Green, red, and blue boxes denote correct predictions, false positives, and ground truth, respectively.
}
  \label{fig:qual}
\end{figure*}
\section{Method}
\label{sec:method}

\subsection{Overview}
As shown in Fig.~\ref{fig:method}, to achieve efficient open-vocabulary cytopathology detection, we propose \textbf{CytoSPM}, a structural prompt matching framework based on a decoupled two-stage design. Inspired by the efficient dual-tower detector WeDetect~\cite{fu2026wedetect}, CytoSPM further leverages the medical textual knowledge from BioMedCLIP~\cite{zhang2023biomedclip} to improve generalization to unseen cell categories. Unlike methods that rely only on class names for region classification, CytoSPM constructs a cytomorphology prompt bank to complement the fine-grained morphological semantics missing from class names.

Specifically, CytoSPM first extracts class-agnostic region-level visual representations through the visual branch, without introducing class names, textual prompts, or cross-modal fusion. Then, in the class-aware stage, class names and structured morphology prompts are matched with these visual regions to dynamically introduce category-relevant morphological knowledge. Through this decoupled design, CytoSPM preserves the efficiency of dual-tower detection while enabling effective recognition of fine-grained open-vocabulary cytopathology categories.

% To achieve efficient open-vocabulary cytopathology detection, we propose
% \textbf{CytoSPM}, a structural prompt matching framework based on a
% decoupled two-stage design. The overall framework is inspired by the
% efficient dual-tower detector WeDetect~\cite{fu2026wedetect} and further leverages the medical
% textual knowledge provided by the vision--language foundation model
% BioMedCLIP~\cite{zhang2023biomedclip} to improve open-vocabulary generalization. Different from
% using only class names for region classification, CytoSPM introduces a
% structured cytomorphology prompt bank to complement the fine-grained
% morphological semantics missing from class names and aligns it with
% class-agnostic visual representations through structural prompt
% matching.

% Specifically, CytoSPM consists of two stages. The first stage is a
% \textbf{class-agnostic visual feature extraction stage}, which extracts
% general region-level visual representations from the input image without
% introducing class names, textual prompts, or cross-modal fusion. The
% second stage is a \textbf{class-aware structural prompt matching stage},
% which performs class-aware matching by combining class names and
% structured morphology prompts with the visual regions generated in the
% first stage. In this way, the visual branch provides reusable visual
% evidence, while the text branch provides class-specific structural
% knowledge through lightweight matching rather than expensive cross-modal
% fusion.

\subsection{Stage I: Class-agnostic Visual Feature Extraction}

Given an input cytopathology image \(I\), the first stage generates
region-level visual representations using only the visual branch.
Specifically, the image is first processed by the
\textbf{visual backbone (ConvNeXt-T)}~\cite{liu2022convnet}, followed by the neck and
detection head to obtain a set of visual region features:

\begin{equation}
V=\{v_i\}_{i=1}^{N}, \quad v_i\in\mathbb{R}^{D},
\end{equation}

where \(v_i\) denotes the feature representation of the \(i\)-th visual
region, \(N\) denotes the number of regions, and \(D\) denotes the
embedding dimension. Since no class names, textual prompts, or
cross-modal fusion are introduced at this stage, the resulting visual
features are class-agnostic and can be reused for matching with arbitrary
categories in the subsequent stage.

\subsection{Stage II: Class-aware Structural Prompt Matching}

The second stage introduces class semantics and cytomorphological
structural knowledge into the detection matching process. For each class
\(c\), we denote its class name as \(y_c\). The class name is encoded by
the frozen BioMedCLIP~\cite{zhang2023biomedclip} text encoder \(f_{\text{text}}(\cdot)\) to obtain
the class text feature:

\begin{equation}
t_c=f_{\text{text}}(y_c), \quad t_c\in\mathbb{R}^{D}.
\end{equation}

However, a class name alone usually provides only coarse-grained
semantics and cannot explicitly describe the structural morphological
features required to distinguish visually similar cytopathology
categories. Therefore, for each class, we construct a structured
cytomorphology prompt bank. Let \(q_j^c\) denote the \(j\)-th morphology
prompt of class \(c\). Each prompt is encoded by the same frozen
BioMedCLIP text encoder:

\begin{equation}
b_j^c=f_{\text{text}}(q_j^c), \quad
B_c=\{b_j^c\}_{j=1}^{M}, \quad b_j^c\in\mathbb{R}^{D},
\end{equation}

where \(b_j^c\) denotes the \(j\)-th morphology prompt feature of class
\(c\), and \(M\) denotes the number of prompts associated with this
class. The prompt bank is stored as a fixed buffer and does not introduce
additional learnable parameters along the class dimension. During
novel-class evaluation, only the class-name embeddings and the
corresponding prompt bank are replaced, while the detector weights remain
unchanged.

\noindent
To connect the class-agnostic visual representations with the
class-aware prompt bank, CytoSPM introduces a structural prompt matching
module. First, the model computes the similarity between each visual
region and each morphology prompt, obtains the maximum response of each
region to the prompt bank, and selects the Top-\(P\) visual regions as
class-related visual evidence:

\begin{equation}
a_i^c = \max_j \left(v_i^{\top} b_j^c\right), \quad
K_v^c = \mathrm{TopP}_{v_i}(a_i^c).
\end{equation}
Here, \(a_i^c\) denotes the maximum response of visual region \(v_i\) to
the morphology prompt bank of class \(c\). \(P\) denotes the Top-\(P\)
visual-region selection budget, and \(K_v^c\) denotes the selected
class-related visual evidence set.

Next, the model uses the selected class-related visual regions to identify
the most relevant morphology prompts. For each prompt \(b_j^c\), its visual
support is computed as:
\begin{equation}
r_j^c = \max_{v_i \in K_v^c} v_i^{\top} b_j^c, \qquad
K_c = \mathrm{TopQ}_{b_j^c}(r_j^c).
\end{equation}
Here, \(r_j^c\) measures the support of the current image for the \(j\)-th
morphology prompt, \(Q\) denotes the Top-\(Q\) prompt selection budget, and
\(K_c\) denotes the selected morphology prompt set. This mutual selection
process keeps only visually supported structural prompts, reducing noise
from irrelevant morphology attributes. The selected structural prompts are then aggregated into an
image-conditioned structural prototype:

\begin{equation}
p_c=\sum_{b_j^c\in K_c}\alpha_j b_j^c,
\end{equation}

where \(p_c\) denotes the structural semantic prototype of class \(c\) in
the current image. The aggregation weight is defined as:\(\alpha_j=\mathrm{softmax}(r_j^c/\tau),\) where \(\tau\) is a temperature parameter. Different from the original
class-name feature \(t_c\), which mainly represents the coarse semantic
meaning of the class name, \(p_c\) represents fine-grained morphological
semantics supported by visual evidence in the current image.

\subsection{Final Matching Score}

For a visual region \(v_i\) and class \(c\), the final matching score is
composed of three complementary matching signals:

\begin{equation}
s(v_i,c)=
v_i^\top t_c
+\lambda_1v_i^\top p_c
+\lambda_2\max_{b_j^c\in K_c}v_i^\top b_j^c.
\end{equation}

The first term is \textbf{class-name matching}, which preserves the
basic semantic information provided by the class name. The second term
is \textbf{structural prototype matching}, which measures the consistency
between the visual region and the image-conditioned structural
prototype. The third term is \textbf{fine-grained attribute matching},
which measures the maximum response between the visual region and the
most relevant morphology prompt of the current class. The final
classification score is jointly determined by these three complementary
matching signals.

\subsection{Warm-up for Structural Matching}

At the early stage of training, region-level visual representations are
not yet stable, and directly introducing structural prompt matching may
bring noisy supervision and affect base-class training. Therefore, we
adopt a structural matching warm-up strategy, allowing the model to first
learn stable basic detection ability from class-name matching and then
gradually introduce structural prototype matching and fine-grained
attribute matching.

Specifically, we define a warm-up coefficient:

\begin{equation}
\rho(t)=\min\left(\frac{t}{T_{\mathrm{warmup}}},1\right),
\end{equation}

and apply it to the two structure-related matching terms:

\begin{equation}
s(v_i,c)=
v_i^\top t_c
+\rho(t)\lambda_1v_i^\top p_c
+\rho(t)\lambda_2\max_{b_j^c\in K_c}v_i^\top b_j^c.
\end{equation}

The class-name matching term always keeps its full weight, while the
structural prototype matching and fine-grained attribute matching terms
are gradually strengthened during training. During inference, we set
\(\rho(t)=1\) and use the complete three-term matching score for
open-vocabulary detection.

% Overall, CytoSPM adopts a decoupled design with class-agnostic visual
% feature extraction and class-aware structural prompt matching. Combined
% with the medical textual knowledge provided by BioMedCLIP, the proposed
% framework improves open-vocabulary recognition of fine-grained
% cytopathology categories while preserving the efficiency of dual-tower
% detection.

\section{Experiments}
\label{sec:exp}

\subsection{Experimental Setup}
\noindent\textbf{Implementation Details.}
\label{sec:exp:impl}
We use ConvNeXt-T as the trainable visual backbone and a frozen BiomedCLIP text encoder to encode class names and structured cytomorphology prompts. The encoded prompts form a fixed prompt bank without introducing additional class-specific learnable parameters. The weights of structural prototype matching and fine-grained attribute matching are set to \(\lambda_1=0.1\) and \(\lambda_2=0.05\), with \(P=0.01\) and \(Q=0.95\). The model is trained for 9 epochs using AdamW with a learning rate of \(3\times 10^{-4}\), weight decay of 0.05, and batch size of 64. Structural matching is warmed up for 4240 iterations. During novel-class evaluation, model weights remain fixed, and only the class-name embeddings and prompt banks are replaced.

\noindent\textbf{Evaluation Datasets and Metrics.}
We evaluate CytoSPM on PentaCyto, which covers five cytopathology domains: respiratory, cervical, thyroid, urinary, and serous fluid cytology. Categories are split into base classes for training and novel classes for zero-shot open-vocabulary evaluation, where only the class-name embeddings and corresponding prompt banks are replaced at test time. We report performance on both base and novel classes using mAP, mAP50, and mAP75.

\subsection{Main Results}
\noindent\textbf{Quantitative Result.}
Table~\ref{tab:main} presents the main comparison results on PentaCyto. For base categories, CytoSPM achieves 33.55 AP, 52.20 AP$_{50}$, and 38.17 AP$_{75}$, obtaining the best AP$_{50}$ among all methods and maintaining strong standard detection performance on seen categories. For novel categories, CytoSPM shows a clearer advantage, achieving 15.00 AP, 21.00 AP$_{50}$, and 17.20 AP$_{75}$, outperforming all baselines on all three metrics. Compared with the strongest baseline YOLO-World-L, CytoSPM improves novel AP, AP$_{50}$, and AP$_{75}$ by 1.30, 2.70, and 1.40, respectively, demonstrating stronger open-vocabulary transfer to unseen cellular categories. Furthermore, CytoSPM achieves the highest AP$_{\mathrm{AVG}}$ of 24.27, surpassing WeDetect with 23.34, indicating a better balance between base-category detection and novel-category generalization. With its decoupled design, CytoSPM avoids expensive deep cross-modal fusion and reaches 25.8 FPS, achieving a favorable trade-off between accuracy and efficiency.

% =====================================================================

% Score (per region v, class c), from region_attribute_fusion.py:
%   s = beta * <v,t_c>            (term1, class-name anchor; always on)
%     + lambda1 * <v,p_c>         (term2, image-conditioned prototype, mutual selection)
%     + lambda2 * max_j <v,b_j^c> (term3, best prompt-bank phrase match)
%   lambda1/lambda2 weight the structural terms.
%   tab:abl_terms -> score-term combination
%       term1 only        -> lambda1=lambda2=0  (== biomedclip baseline)
%       term1 + term2     -> struct_lambda2_fixed=0
%       term1 + term3     -> struct_use_term2=False
%       full (term1+2+3)  -> shaded row
%   tab:abl_pq    -> JOINT 2-D sweep, Top-Q (rows) x Top-P (cols).
%       Top-P = PERCENT of regions pre-selected per feature level (~1%).
%       Top-Q = PERCENT of prompt-bank phrases kept per class (0<q<1; code
%               struct_topq fraction path); main Q=0 keeps all phrases.
% Banks: data/texts/tct_ngc_descriptors_*_noabs_per_attr_biomedclip.pth.
% =====================================================================

% ----- (A) score-term combination -----------------------------------
% Keep Table IV and Fig. 5 together, with independent captions and counters.
\begin{figure}[!t]
  \centering
  \begin{minipage}{\linewidth}
  \centering
  \small
  \captionof{table}{Ablation of the three matching signals in the final matching score
on the PentaCyto. t1 denotes class-name matching, t2 denotes
structural prototype matching, and t3 denotes fine-grained attribute matching.
The best result in each metric row is shown in \textbf{bold}.}
  \label{tab:abl_terms}
  \setlength{\tabcolsep}{5.5pt}
  \begin{tabular}{l l ccc}
    \toprule
    Split & Terms & AP & AP$_{50}$ & AP$_{75}$ \\
    \midrule
    Base  & t1+t2+t3 & 33.55 & 52.20 & \textbf{38.17} \\
          & t1+t2    & \textbf{33.67} & \textbf{52.37} & 37.99 \\
          & t1+t3    & 33.08 & 51.35 & 37.61 \\
    \midrule
    Novel & t1+t2+t3 & \textbf{15.00} & \textbf{21.00} & \textbf{17.20} \\
          & t1+t2    & 14.80 & 20.90 & 17.00 \\
          & t1+t3    & 13.90 & 20.40 & 16.70 \\
    \bottomrule
  \end{tabular}
  \end{minipage}
  \par\vspace{\floatsep}
  \includegraphics[width=\linewidth]{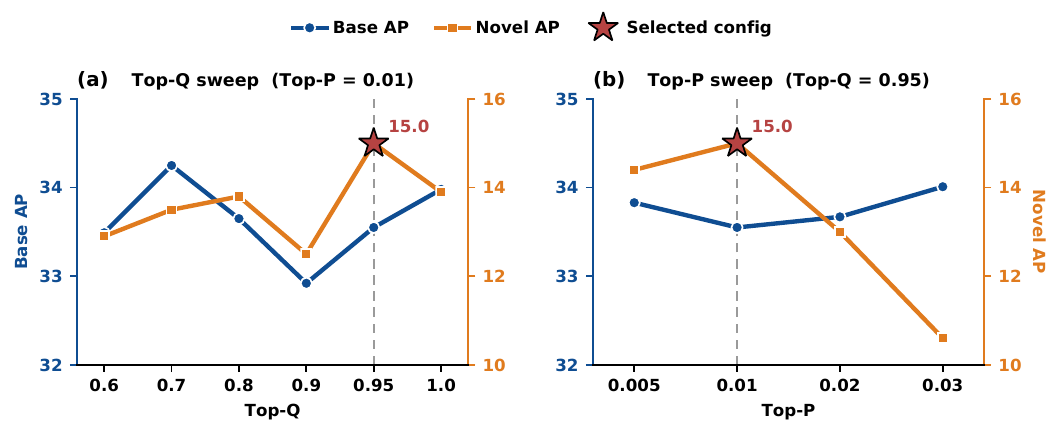}
  \caption{
\textbf{Ablation of Top-$Q$ and Top-$P$.}
Top-$Q$ selects morphology prompts, while Top-$P$ selects visual evidence for structural matching.
}
  \label{fig:topqp_ablation_wide}
\end{figure}
% Keep warm-up above text-encoder ablation within a single column.
\begin{figure}[!t]
  \centering
  \includegraphics[width=\linewidth]{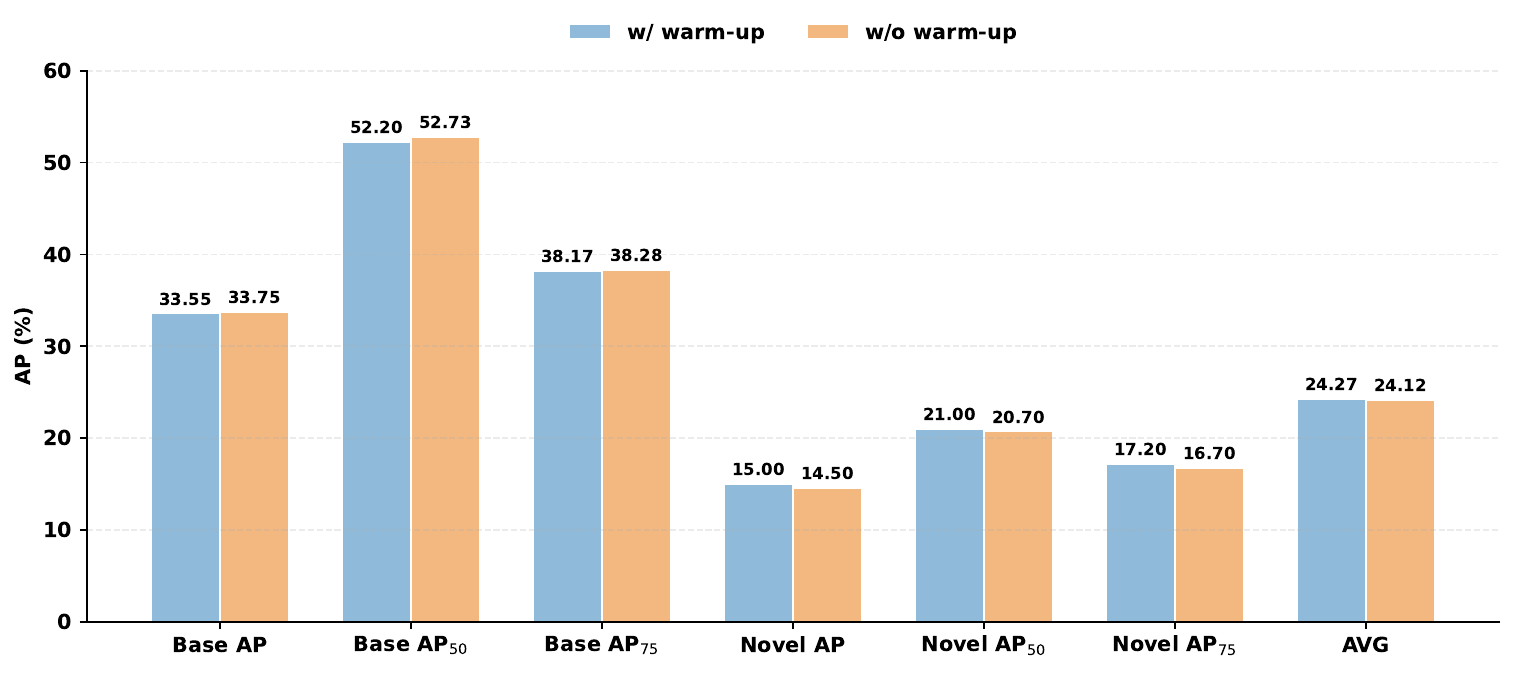}
  \caption{Effect of the structural matching warm-up in CytoSPM.}
  \label{fig:abl_w_rmup}
  \par\vspace{\floatsep}
  \begin{minipage}[t]{0.48\columnwidth}
    \centering
    \includegraphics[width=\linewidth]{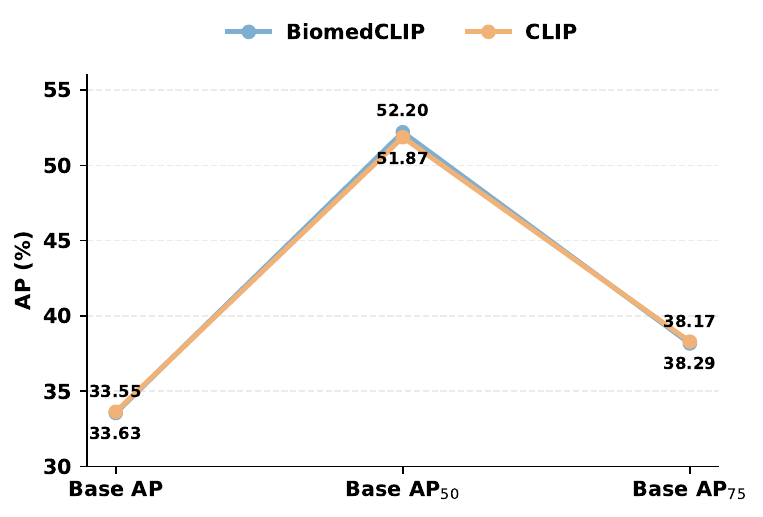}
  \end{minipage}\hfill
  \begin{minipage}[t]{0.48\columnwidth}
    \centering
    \includegraphics[width=\linewidth]{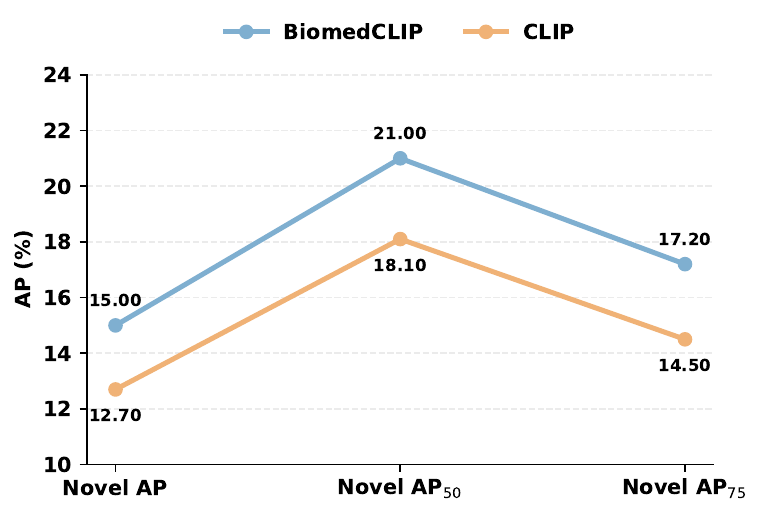}
  \end{minipage}
  \caption{Effect of the text encoder in CytoSPM on the PentaCyto. BiomedCLIP and CLIP show comparable performance on base categories, while BiomedCLIP achieves clearly better results on novel categories, indicating the benefit of domain-specific medical text representations for open-vocabulary cytopathology detection.}
  \label{fig:abl_text_encoder}
    \label{fig:abl_text_encoder_base}
    \label{fig:abl_text_encoder_novel}
\end{figure}

\noindent\textbf{Qualitative Results.}
\label{sec:exp:qual}
Figure~\ref{fig:qual} compares CytoSPM with representative open-vocabulary detectors across different cytopathology scenarios. CytoSPM localizes target cells more accurately and produces fewer false positives under challenging conditions, including cluttered backgrounds, morphologically similar cells, and held-out novel categories. For base categories such as lymphocyte and SHGUC, CytoSPM better matches the ground truth, while WeDetect and YOLO-World-L tend to miss targets or confuse categories. For the novel thyroid malignant category, CytoSPM still localizes the malignant cell region, showing stronger transfer of cytomorphological knowledge to unseen categories.

\subsection{Ablation Studies}
\label{sec:exp:ablation}
\noindent\textbf{Ablation of Matching Signals.}
Table~\ref{tab:abl_terms} analyzes the contribution of the three matching
signals in the final matching score. Using class-name matching and
structural prototype matching already provides strong base-category
performance, showing that the image-conditioned structural prototype can
effectively complement the class-name anchor. Adding fine-grained
attribute matching further improves novel-category detection, achieving
the best Novel AP, AP$_{50}$, and AP$_{75}$. This indicates that directly
matching visual regions with the most relevant morphology prompts helps
the model capture fine-grained cytomorphological evidence and improves
open-vocabulary transfer to held-out categories.

\noindent\textbf{Ablation of Top-\(P\) and Top-\(Q\).}
Figure~\ref{fig:topqp_ablation_wide} analyzes the effects of Top-\(P\) region selection
and Top-\(Q\) prompt selection in CytoSPM. For Top-\(Q\), keeping all prompts
(\(Q=1\)) gives slightly stronger base-category performance, but does not
produce the best novel-category results, suggesting that the full prompt bank
may contain morphology attributes that are less relevant to unseen categories.
With a moderate prompt selection ratio, \(Q=0.95\) achieves the best Novel AP,
AP$_{50}$, and AP$_{75}$. For Top-\(P\), selecting too few or too many visual
regions weakens the alignment between visual evidence and morphology prompts.
The best novel performance is obtained at \(P=0.01\), showing that a compact
set of highly relevant visual regions provides more reliable evidence for
structural prompt matching. These results support the mutual-selection design
of CytoSPM, where visually supported regions and morphology prompts are
dynamically selected to reduce noise and improve open-vocabulary transfer.

\noindent\textbf{Warm-up Ablation.}
Figure~\ref{fig:abl_w_rmup} analyzes the effect of structural matching warm-up in CytoSPM. Without warm-up, the model performs slightly better on base categories, but warm-up improves novel AP from 14.50 to 15.00 and also raises AP$_{50}$ and AP$_{75}$ to 21.00 and 17.20. This suggests that gradually introducing structural matching helps avoid unstable early supervision and improves open-vocabulary generalization to unseen categories.

\noindent\textbf{Text Encoder Ablation.}
Figure~\ref{fig:abl_text_encoder} analyzes the influence of the text encoder
in CytoSPM. BiomedCLIP and CLIP obtain similar performance on base categories,
indicating that both text encoders can provide effective class-name guidance
for seen categories. However, BiomedCLIP consistently outperforms CLIP on
novel categories, especially on Novel AP$_{50}$ and AP$_{75}$. This suggests that BiomedCLIP is more suitable as the text encoder for CytoSPM, since it is specifically designed for medical applications and trained on large-scale medical image--text pairs, making it better suited for medical image understanding and open-vocabulary cytopathology detection.

\section{Conclusion and Limitations}
\label{sec:concl_limit}

We present \textbf{PentaCyto}, a multi-domain benchmark for open-vocabulary cytopathology detection, and \textbf{CytoSPM}, an efficient detector based on a decoupled two-stage design. PentaCyto provides a unified evaluation protocol across five cytology domains. CytoSPM first extracts reusable class-agnostic visual representations, and then performs class-aware structural prompt matching with class names and cytomorphology prompts. By combining class-name matching, structural prototype matching, and fine-grained attribute matching, CytoSPM improves novel-category recognition while avoiding heavy cross-modal fusion. However, CytoSPM still relies on biomedical vision--language pretraining, and its generalization to broader microscopy scenarios remains to be further studied.
% \input{section/supplement}   % optional supplementary material

% =====================================================================
% \section*{Acknowledgment}
% \todo{Funding, GPU time, dataset providers.}

% =====================================================================
\bibliographystyle{IEEEtran}
\bibliography{paper}

\end{document}